# Analyzing language development from a network approach

Jinyun KE
English Language Institute
University of Michigan, Ann Arbor
jyke@umich.edu

Yao YAO
Department of Linguistics
University of California, Berkeley
yaoyao@berkeley.edu

Abstract

In this paper we propose some new measures of language development using network analyses, which is inspired by the recent surge of interests in network studies of many real-world systems. Children's and care-takers' speech data from a longitudinal study are represented as a series of networks, word forms being taken as nodes and collocation of words as links. Measures on the properties of the networks, such as size, connectivity, hub and authority analyses, etc., allow us to make quantitative comparison so as to reveal different paths of development. For example, the asynchrony of development in network size and average degree suggests that children cannot be simply classified as early talkers or late talkers by one or two measures. Children follow different paths in a multi-dimensional space. They may develop faster in one dimension but slower in another dimension. The network approach requires little preprocessing of words and analyses on sentence structures, and the characteristics of words and their usage emerge from the network and are independent of any grammatical presumptions. We show that the change of the two articles "the" and "a" in their roles as important nodes in the network reflects the progress of children's syntactic development: the two articles often start in children's networks as hubs and later shift to authorities, while they are authorities constantly in the adult's networks. The network analyses provide a new approach to study language development, and at the same time language development also presents a rich area for network theories to explore.

## 1. Introduction

Children acquire their language in different ways. Various kinds of measure have been used to compare the trajectories of development. Among them, vocabulary size and Mean Length of Utterance (MLU) are two basic ones used frequently for evaluating the rate of early development. Two typical styles of learners have been identified: early vs. late talkers (Bates & Goodman, 1997). Early talkers tend to have a large vocabulary size and long sentences (i.e. a large MLU) earlier than other children, and late talkers lag behind the average. Children also vary in the distribution of types of words in their early vocabulary and the order of acquisition. For example, some children acquire more nouns for objects at the early stage, while some use more formulaic expressions, such as "Lemme see", "Don't do dat", in their speech. This contrast is termed as a dichotomy between "referential" and "expressive" style (Nelson, 1973). Different types of learning styles have also been identified in word learning and phonological development. (Bates et al., 1988; Shore, 1995).

Research has shown significant differences in the development in morphology and syntax as well. For example, children differ in the rates and routes of acquisition of different grammatical morphemes, such as the pluralistic function of "-s" in English (Brown 1973), as well as various grammatical constructions, such as auxiliaries (Stromswold, 1990), questions (Stromswold, 1995), and so on. Very often these measures of differences are based on text analyses of corpora collected from spontaneous speech. It is often presumed that the appearance of construction in children's speech data indicates the existence of the knowledge in the children corresponding to the grammars in the adults. However, such presumption need to be taken with caution, as children use certain constructions often as formulaic expressions, instead of knowing the constructions are decomposable as adults do (Peters, 1977; Wray, 2002).

In this paper, we propose a novel approach to analyze the language development from a network perspective. We take the children's longitudinal speech data to construct lexical networks at different stages, and compare these networks with several measures. Also, we compare networks between different children, and networks between children and their caretakers. Through the comparison, children's development and individual differences can be demonstrated quantitatively by measures with little language-specific assumptions.

Networks have been extensively studied within the area of graph theory in mathematics. Random networks used to be the main object of research and were often assumed as the default model for real-world networks. However, recent studies on several real-world networks, such as collaboration networks, the WWW and Internet, have revealed some interesting features, which cannot be captured by random networks (Watts & Strogatz, 1998; Barabási & Albert, 1999). It has been found that many complex networks in the real world are like "a *small world*" (Watts & Strogatz, 1998): regardless of the large size of the network, any two nodes in the network can be connected through only a small number of intermediate nodes, and directly connected nodes often share common neighbors (resulting in a high "cluster coefficient" of the network). Meanwhile, studies have shown that in many networks, there are a number of nodes having an extremely large number of connections while most nodes only have a handful; the existence of the so-called "hubs" make the network appear to be *scale-free* (Barabási & Albert, 1999).

The discovery of these features has triggered a new surge of network research in recent years (some general accounts of the development of the field can be found in Barabási, 2002 and Buchanan, 2002). There is an increasing interest in networks in a broad range of disciplines. It has been shown that networks of various natures share the small-world and/or scale-free properties, which implies the existence of some universal principles for network formation and evolution in the real world. Moreover, when real-world systems are reformulated from a network perspective, we may obtain new insights on old questions which traditional approaches are not able to achieve. For example, as the social networks are found to be scale-free networks, the traditional public health approach of random immunization could easily fail. Instead identifying and immunizing the hubs in the network may provide a more effective way to stop epidemics (Barabási & Bonabeau, 2003).

The development in network research has inspired interests in networks in language. Though the lexicon had been considered as a network (Aitchison, 1994), this was mostly taken as a metaphoric concept and there had been few rigorous analyses until recently. In the last few years, researchers started to construct and analyze lexical networks in various ways (e.g. Ferrer & Solé, 2001; Dorogovtsev & Mendes, 2001; Motter et al, 2002; Sigman & Cecchi 2002; Ferrer et al, 2004, Steyvers & Tenenbaum 2005). For instance, Ferrer & Solé (2001) use the British National Corpus to construct gigantic word networks (with the size of over 460,000 nodes). The network is built based on collocation relationship: words are connected if they are direct neighbors in a sentence. The resultant network exhibits some small-world characteristics. Motter et al. (2002) report similar results, though their lexical networks are constructed in a different way: the words are connected by the synonym relationship given by an English thesaurus. The compact structure of the networks may provide some insights for explaining the high speed in online language processing. Sigman & Cecchi (2002) examine lexical networks of English nouns. Words are connected according to various semantic relationships provided by WordNet, such as synonymy, antonymy, hyponymy/ hypernymy, meronymy/holonymy and polysemy. They find that the presence of the relationship of polysemy dramatically changes the organization of the network to be more compact.

Inspired by these studies on networks in language, we are interested in applying network analyses to language development. While earlier studies all deal with static networks, we will examine lexical networks from a dynamic perspective, by comparing networks constructed from longitudinal data. In this paper, we report two sets of measures on the networks. The first set of measure consists of two independent parameters for comparing the growth of the networks, that is, the size and the connectivity. The second measure relies on the analysis of hub and authority nodes in the networks. These measures on the networks provide some new methods to compare children's lexical and syntactical development.

## 2. The material and the methodology

The corpus used in this study is the Manchester corpus from CHILDES (Theakston et al., 2001). The corpus consists of a year-long study of twelve English-speaking children, six boys and six girls. At the beginning of the study, the children ranged in age from 1;8.22 to 2;0.25, with MLUs ranging from 1.06 to 2.27. Each child was visited at home twice every three weeks, and in total has 34 visits. The duration of recording for each visit was one hour, divided into two parts with a 10-minute break in between. During the visits, children were engaged in normal play activities with their mothers.

The spontaneous utterances from the children and their mothers are extracted from the transcripts and used as input to their respective networks. The word forms in the utterances are taken as individual nodes, and the collocation relationships between the words are taken as the link between the nodes. The links are directional, that is, if word A appears before word B in an utterance, there will be a link from word A to word B. To give an example, Figure 1 shows a part of the conversation between a mother and a child, and two constructed networks, one for the mother, and the other for the child, are displayed in Figure 2. For instance, there are directed links between "what" and "would" in the mother's network, between "and" and "lion" in the child's network. And there are a number of isolated nodes, especially in the children's networks.

```
*MOT: what would you like for your birthday ?
*MOT: would you like a train ?
*MOT: Joel ?
*CHI: yes .
*CHI: oh .
*CHI: I'd like a elephant .
*CHI: no .
*CHI: and lion .
```

Figure 1. An example of conversation between a mother and a child in the Manchester corpus.

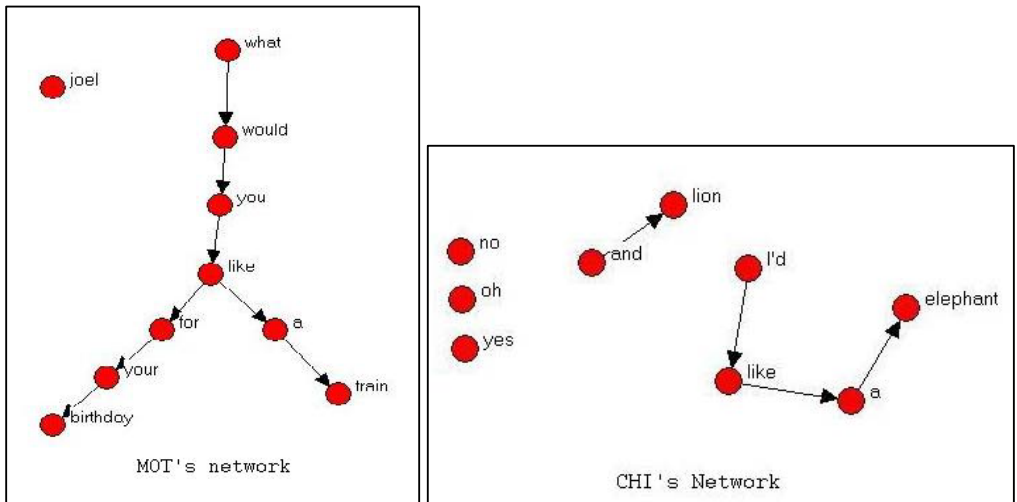

Figure 2. The mother's and child's networks constructed from the conversation given in Figure 1. The network diagrams are drawn using the free software **Pajek 1.00**, which is downloadable from http://vlado.fmf.uni-lj.si/pub/networks/pajek/.

In the transcripts of the corpus, non-spontaneous speech, including imitation, self repetition, routines and utterances partly intelligible, have been marked. These data are excluded when constructing the networks, assuming this would provide a better picture of the actual production capacity of the child.

In our network model, a node is a unique word form, rather than a lemma. The word forms "bus" and "buses", "go" and "goes" are all represented by different nodes. In doing so, we avoid the problem of determining whether the conjugation or derivative forms produced by the young children are learned independently as individual lexical items, or derived from productive morphological rules.

We have developed two ways to build the networks. One is to build the networks by accumulating the data from the beginning, and the other builds independent networks for different stages defined by the MLU. These two types of networks are called accumulative networks and staged networks, respectively. The staged networks provide a more revealing picture of the development, to be discussed in Section 3.2. Therefore, after performing the analyses of network growth on both types of networks, we only apply the analyses of hubs and authorities to the staged networks. In the following, we will introduce these networks and report the results of the analyses.

### 3. Analysis 1: network growth

#### 3.1  Accumulative networks

First, we examine children's development by comparing the networks constructed accumulatively over time. A network for a child at a time instant $t$ is built from sentences produced by the child in the visit of time $t$ and all the proceeding visits before $t$. Going through the data from the 34 visits of each of the 12 children in the corpus, we construct 34 networks for each child and perform two simple measures on these networks.

The first measure is the *size* of the network, $N$, that is, the number of nodes in the network. It corresponds to the measure of vocabulary size, which is often taken as an index for the rate of lexical development in traditional studies. Figure 3 shows the sizes of the 34 accumulative networks of each of the 12 children at 34 time instants. Except for Ruth, most children exhibit a relatively linear growth during the time span of investigation, and different children have different growth rates. Ruth has a slow increase at the beginning period and then shows a "burst", i.e. a rapid increase of her vocabulary (Bates & Goodman, 1997). Other children seem to have all passed this period at the time when the investigation started. Among them, Joel has the fastest increasing rate in vocabulary size, starting at about reaching about 2300 words at the time of 2;10.11. In a conventional view, Joel is often considered as an early speaker, and Ruth as a late speaker. However, such a categorization becomes questionable when we go on to compare the second measure.

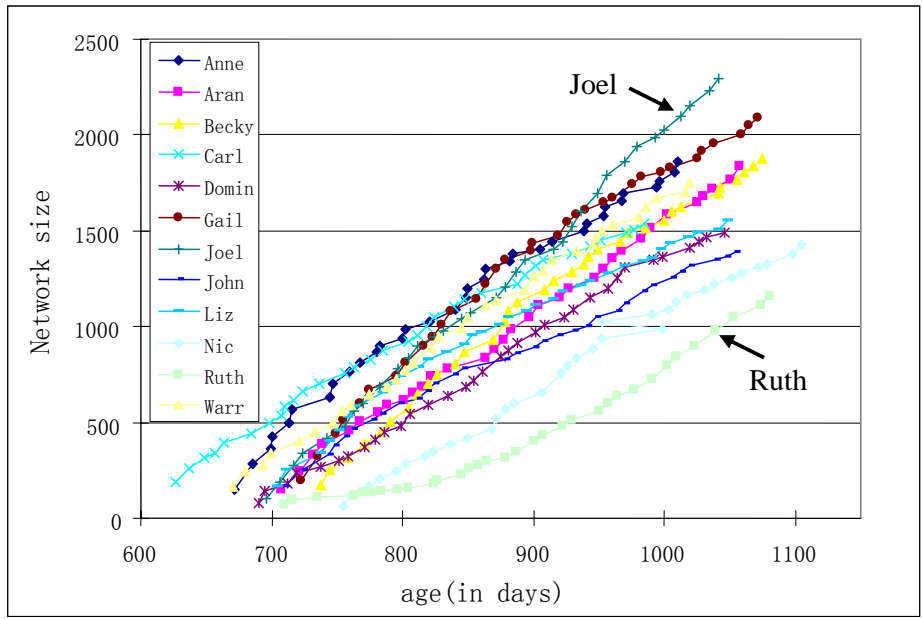

Figure 3. Growth of networks in size over time in the 12 children.

The second measure is the ***average degree*** of the networks, which is calculated by the total number of links divided by the number of nodes. The ***degree of a node*** is the number of links it possesses. The average degree is taken as an index for the connectivity or density of a network. Most of the large-size networks in reality are sparse networks, i.e. the average degree is significantly small with respect to the size of the networks. So far, there is no available measure in traditional linguistic analyses which corresponds to this variable of average degree of the lexical network. In fact, the larger degree of a node implies the higher degree of variation of combination the word has with other words. Thus we can interpret the average degree as an integrated index of the productivity of the words in use and of the complexity of the utterances.

In the constructed networks, the links have directions, and thus there are two kinds of degrees, ***in-degree*** and ***out-degree***. The in-degree of a node measures the number of links from other nodes to this node, while the out-degree is the number of links going from this node to others. Obviously, in one network, the sum of in-degrees and that of out-degrees are identical, and hence the average in-degree and out-degree are the same. So here we only calculate the average out-degrees. Figure 4 shows the average out-degrees of the 12 children's accumulative networks.

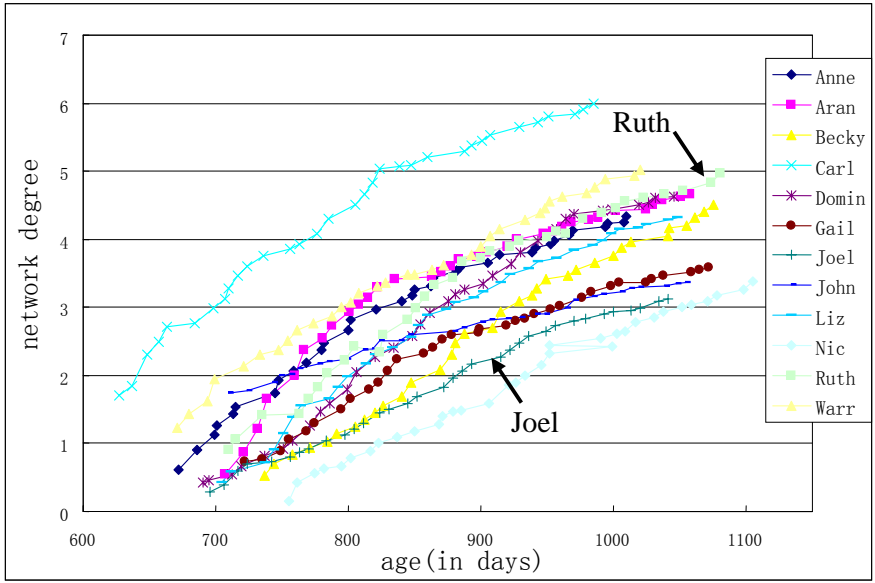

Figure 4. Growth of networks in connectivity (i.e. average degree) over time in the 12 children.

Similar to the network size, the average degree in the networks also grows linearly over time in most children. However, when we compare the sizes and the average degrees of different children's networks, there is little correlation between the two measures. This can be seen in the scatter-gram of the size and average degree of the last accumulative networks of the 12 children, as shown in Figure 5. Most children cluster in the middle area, which refers to networks with a medium network size (around 1400-2200 words) as well as a medium average degree (around 3-5 links per word).

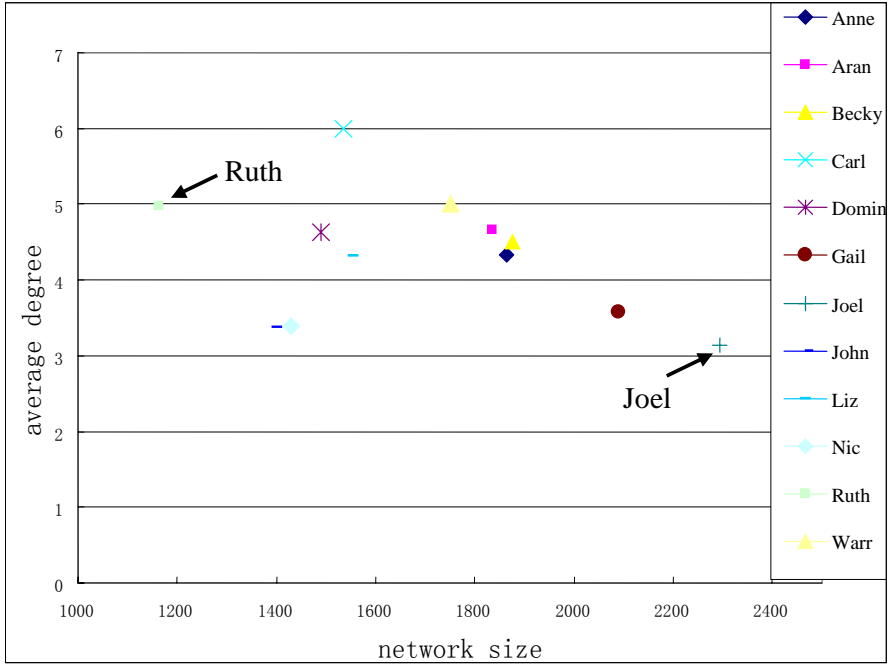

Figure 5. Size vs. average degree of the last accumulative networks of the 12 children.

The two children, Joel and Ruth, who exist at the two extremes of vocabulary size, also exhibit a clear distinction in average degree, but in a reverse opposition. Joel's network size is almost twice as large as that of Ruth's, while Ruth's average degree exceeds that of Joel's by about 70%. With respect to vocabulary size, Ruth may be considered as a late speaker and Joel as an early speaker. However, their difference in average degree suggests an opposite dichotomy of development. Despite starting from a similar average degree, Ruth has a much faster development in word usage than Joel, as shown in Figure 4. This suggests that some children having a smaller vocabulary exploit the known words more flexibly and productively, which is intuitively reasonable. There is one outlier in the development of average degree - Carl, who has a much higher average degree (around 6 links per node in the last network) than other children, while his vocabulary size is within the medium range.

From the distribution of the 12 children shown in Figure 5, we can see that the two parameters, i.e. vocabulary size and average degree, are mostly independent from each other. To evaluate the rate of language development, we should not only pay attention to the number of words the children acquire, but also how they exploit these words. From this result, we can see that the network representation of the data and the analyses performed on the networks provide us with some new measures, which may not be obtained easily with traditional text analyses.

### 3.2 Stage Networks

The above accumulative networks preserve all prior utterances and hence are less vulnerable to low sampling frequency in the corpus. It has been shown that sampling frequency may affect the analyses in studying the development of different linguistic features (Tomasello & Stahl, 2004). However, the accumulative network model assumes that children's language development is purely an additive process, as new nodes and links are continuously added to the network and will never be removed. This assumption is not realistic. During the course of language development, children constantly revise their language system. Early child forms like "choochoo" and "nana" gradually fade out; words that are picked up months ago may be forgotten, etc. Also, earlier mistakes are often corrected later. A well-known example is the correction of over-generalization in learning past tense in English. For instance, the form "goed" disappears after the regular past tense morpheme is acquired (Pinker & Ullman, 2002). These changes are all non-additive in nature and cannot be reflected in the accumulative network model.

To overcome the weaknesses of accumulative networks, we propose another type of network model called stage network. The entire time period is divided into stages according to the development of MLU, and a network is constructed for each stage by taking input within the stage.

The reason to choose MLU as the variable to define stages is because MLU is often taken as an effective measure for early language development, in particular for syntactic development. Also, in our corpus, we observe in several children that MLU increases in a step-wise manner: it first fluctuates in a relatively narrow range for a while, and jumps to a higher level, and then stays there for a while, and so on.

Here we report the results of the stage networks for four children. First we include the two extreme children we mentioned earlier, Joel and Ruth, who have the largest and smallest vocabulary size in the last network, respectively. Another child included is Carl who has the largest average degree in the last network. The fourth child is Anne whose vocabulary size and average degree both lie in the middle range.

We divide the MLU into five ranges, [1,1.5] , (1.5, 2 ] , (2, 2.5], (2.5, 3], and (3,3.5] ("[" means including the value and "(" excluding), and separate the time period into five corresponding stages. To ensure that the networks across different stages and across different children are comparable, we try to keep the time span for each network a constant value applied to all children. If an MLU stage is particularly long, we divide it further into an early stage and a late stage, and construct one network for each sub-stage.

As shown in Figure 6, Joel and Ruth are quite close in MLU development, except that Joel stays longer in stage 4 and doesn't really reached stage 5, and Ruth stays longer in stage 3. Hence we split Joel's stage 4 into early stage 4 and late stage 4, and Ruth's stage 3 into early stage 3 and late stage 3. Carl and Anne start with a higher MLU than the other two children, and their growth shows a stronger step-wise tendency.

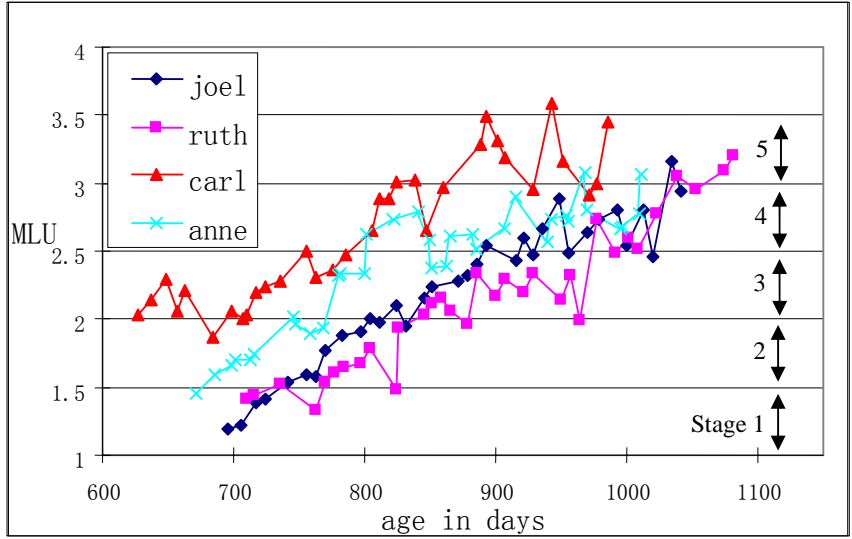

Figure 6. MLU development of four typical children.

Next we pick out five continuous transcription files that cover a time span of around 40 days in each stage/sub-stage (except for Ruth's stage 1 and stage 5, which have only 4 time instants within the given MLU range), and construct networks based on the transcripts corresponding to each stage/sub-stage. Table 1 gives the stages defined for the four children, and lists the information of the data used for network construction, including the number of transcription files, the number of utterances, the number of morpheme tokens for each stage, and the MLU. Having the stage networks constructed for the four children, we first perform the size and average degree analyses on the networks. The results are given in Table 1, and also displayed in Figures 7 and 8.

Table 1. The information used for constructing stage networks, and the sizes and average degrees of the networks for the four children.

|  | Stages | No.of files | no.of utterances | no.of morpheme tokens | MLU | network size | average degree |
|---|---|---|---|---|---|---|---|
| Joel | S1 | 5 | 1091 | 1464 | 1.342 | 422 | 0.7 |
|  | S2 | 5 | 1369 | 2622 | 1.915 | 591 | 1.33 |
|  | S3 | 5 | 2002 | 4588 | 2.292 | 744 | 2 |
|  | early S4 | 5 | 3084 | 8067 | 2.616 | 1060 | 2.36 |
|  | late S4 | 5 | 2404 | 6444 | 2.681 | 875 | 2.35 |
| Ruth | S1 | 4 | 1186 | 1690 | 1.425 | 121 | 1.43 |
|  | S2 | 5 | 1599 | 2613 | 1.634 | 119 | 2.29 |
|  | early S3 | 5 | 1876 | 3981 | 2.122 | 250 | 3.396 |

|      | late S3   | 5 | 1881 | 4179  | 2.222 | 409 | 3.96 |
|------|-----------|---|------|-------|-------|-----|------|
|      | S4        | 5 | 2618 | 6838  | 2.612 | 573 | 3.4  |
|      | S5        | 4 | 2351 | 7227  | 3.074 | 690 | 3.4  |
| Carl | early S3  | 5 | 2481 | 5351  | 2.157 | 398 | 2.71 |
|      | late S3   | 5 | 2554 | 5884  | 2.304 | 509 | 3.08 |
|      | S4        | 5 | 3047 | 8706  | 2.857 | 738 | 3.41 |
|      | early S5  | 5 | 3279 | 10610 | 3.236 | 742 | 3.62 |
|      | late S5   | 5 | 2892 | 9373  | 3.241 | 726 | 3.61 |
| Anne | S2        | 5 | 2070 | 3413  | 1.649 | 525 | 1.44 |
|      | S3        | 5 | 2138 | 4426  | 2.070 | 596 | 2.26 |
|      | early S4  | 5 | 2200 | 5645  | 2.566 | 717 | 2.66 |
|      | middle S4 | 5 | 2146 | 5855  | 2.728 | 755 | 2.73 |
|      | late S4   | 5 | 2048 | 5770  | 2.817 | 771 | 2.80 |

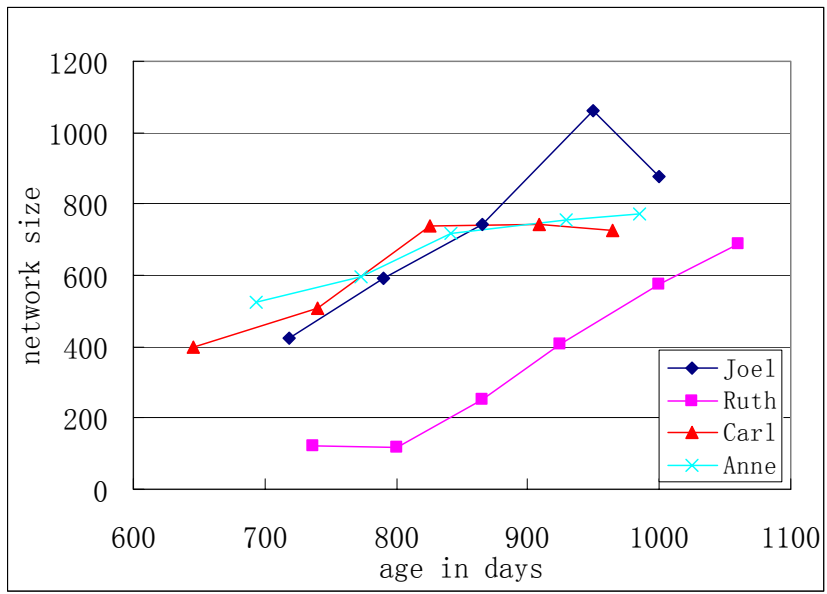

Figure 7. Growth in size of the stage networks of the four children.

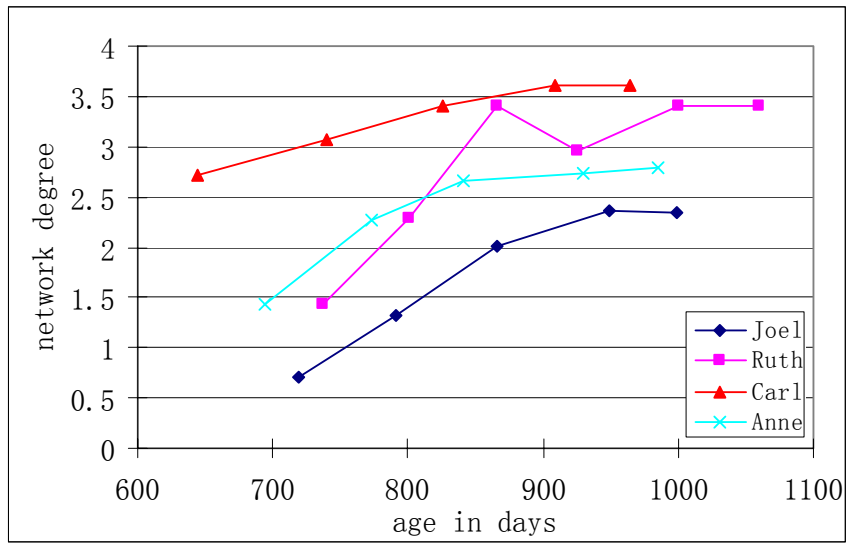

Figure 8. Growth in connectivity (average degree) of the stage networks of the four children.

Figures 7 and 8 show that, similar to the accumulative networks, the four children's stage networks are mostly increasing in both size and average degree over time. In other words, their networks are growing larger and denser simultaneously. While the increase of size of the accumulative networks may be an artifact of the additive network construction, the increase of size in stage networks is not trivial, and reflects a genuine growth in vocabulary in the children. This is clear when we compare stage networks, such as those of stage 4 and 5 for Ruth, and middle and late stage 4 for Anne. These stage networks have fewer utterances, or even fewer tokens of morphemes (in the case of Anne), but they have a larger network size. Moreover, though the growth of networks in average degree is expected, the growth of the stage networks is no more as linear as those in the accumulative networks.

Different children show different rates of growth in these two measures. Ruth starts with the smallest vocabulary among the four children, and after a stagnant period, her vocabulary grows abruptly, featuring a lexical spurt. Meanwhile, its degree has little change. In comparison, Joel starts with a much larger vocabulary, which keeps increasing sharply. However, Joel's network degree is always smaller than that of Ruth's. But this does not mean that a larger network size always accompanies a smaller network degree. The data from Carl and Anne suggest that some children may have both a large network size and a large network degree.

When we plot the size and average degree of the four children's stage networks on one graph, we can see more clearly that children's networks develop along different paths, as shown in the left part in Figure 9. Children give different priorities to the development of network size and density. Despite the distinctive developmental paths, it seems that they are moving toward a similar target in the upper right corner of the graph. To have a better idea where the target for development lies in, we analyze the stage networks of the mothers, to see if the children' networks are approaching those of their mothers'.

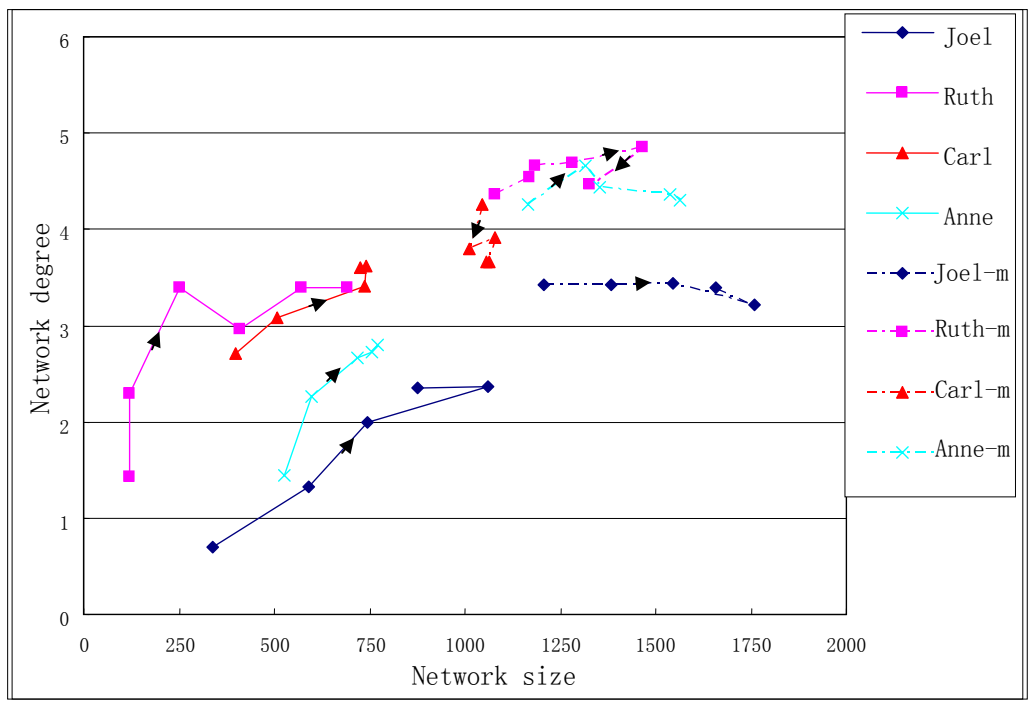

Figure 9. Size vs. average degree of the stage networks of the four children and their mothers.

The stage networks of the mothers all locate in the right upper corner in Figure 9, with larger vocabulary sizes and large average degrees. These networks are the targets of the children's development. The mothers' networks have fewer changes over time than the children's, showing a higher stability. Especially, unlike children's networks growing in both dimensions, the mothers' stage networks show little change in the average degree, and sometimes there is even a decrease in some stages. In particular, Carl's mother's networks exhibit a significant decrease in the average degree, from 4.3 to 3.7. Moreover, the networks of Carl's mother are different from those of other mothers in having little growth in size, which suggests that she speaks in a relatively more stable manner, regardless of the development of child.

The networks of the other three mothers grow in size over time, along with the growth of the children's networks. What we show here are consistent with the findings from research on care-takers' speech: adults adopt different styles when talking to their children (Snow & Ferguson, 1977). Some adults are more responsive to their children and their speech is more adjusted to be child-friendly, while some tend to maintain their normal speech style and have little adjustment when they talk to their children. Carl's mother may belong to the second type, while the other three mothers belong to the first one.

There exist certain correlations between the mothers' and the children's speech in the other three pairs. While Joel leads among the other three children in the development of network size but with the smallest network degree, his mother's networks exhibit similar characteristics, i.e. a large network size but a small network degree. Ruth and her mother show the opposite characteristics in the same way, i.e. a small network size but a large network degree. And the networks of Ann and her mother's lie in between the above two. To explain this correlation, we are confronted with the long-standing chicken-and-egg puzzle of the relationship between children and their care-takers' speech: is it that a child learns faster because his caretaker gives more input, or that the caretaker gives more input because the child shows more interests to learn and is more responsive? We are not able to provide an answer to this puzzle with the data of the three mother-child pairs. However, the disassociation we see in Carl and his mother suggests that some children are predisposed to be able to learn faster, without requiring more input. In other words, in some cases, the chicken-and-egg problem doesn't exist.

## 4. Analysis 2: The shift of the articles

### 4.1 Concepts of hub and authority

The network size and network average degree investigated above are two very fundamental and simple measures of the network characteristics. They only provide some coarse global measures, and the internal structure of the networks and the features of the nodes are not taken into account. In the following, we carry out a set of analyses regarding the detailed network structure, i.e. the hub-authority analysis. The aim is to identify the "important" nodes in the networks. By comparing the changes of the important nodes in the network, we can obtain a quantitative indicator regarding how the structures of the networks change. And by comparing the networks of the children's with their mothers', we may see how the children's networks grow towards the adults', in addition to what the network size and average degree have shown above.

Networks are composed by a set of nodes, but the nodes are not equally important – some are more important than others. In a social network, important nodes may represent leaders and the others followers; in a network of websites connected by hyperlinks, popular websites such as Yahoo and Google may be more important than other sites.

For a directed network, there are two kinds of important nodes – hubs (which point to others) and authorities (which are pointed to). Intuitively, the nodes having more links should be more important in the network. Hubs would be the nodes having the largest out-degrees, and the authorities the largest in-degrees. However, it turns out that it is not the case; the number of connections can't indicate the real importance of nodes. A node which is pointed by a large number of unimportant nodes will be less authoritative than a node which is pointed by a few important nodes (Kleinberg 1998). Similarly, a node will be less "hub"-like, if it points to a large number of unimportant nodes, than that pointing to a few highly authoritative nodes. In other words, a node is only a good hub if it connects with many good authorities, and it is a good authority if it is linked with many good hubs. Hubs and authorities are in fact interdependent with each other, and cannot be identified independently.

It needs some algorithm to determine which nodes are hubs and authorities. Each node has a hub and an authority weight, and these weights are computed with some iterative process. The nodes having higher hub weights are considered as better hubs, and those having higher authority weights are better authorities. A node can be both a good hub and a good authority. The idea of hub and authority has been very useful for internet search engines to identify important and relevant webpages (Kleinberg, 1998; Brin & Page, 1998). For example, the effective algorithm used in Google, a well-known search engine, is developed mainly based on this idea of hub and authority (c.f. http://www.google.com/technology/). In our study, the computation is performed with the free software Pajek (2000) which implements an algorithm similar to the Google's.

Figure 10 gives an example of a small network with 13 nodes to illustrate the existence of hubs. In the figure, node 'a' connects to three nodes: 'b', 'c', and 'd'; node 'h' connects to four nodes: 'k', 'l', 'm' and 'n'. However, 'b', 'c' and 'd' are better authorities than 'k', 'l', 'm', and 'n'. The computed hub weight of 'a' is 0.62, ranking the highest among all the nodes, whereas the hub weight of 'h' is only 0.0002, ranking the fifth. This shows that though 'h' has a larger out-degree than 'a', it is a less important hub than 'a'.

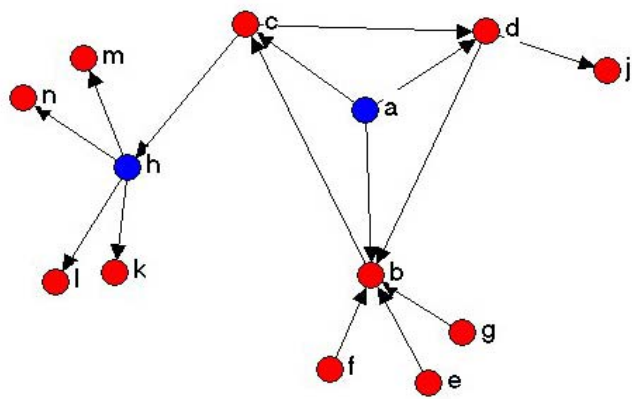

Figure 10. An example of "hubs" in a network.

In the lexical network, the meaning of the hubs and authorities is less obvious than in the case of WWW network. The directed links between nodes in the network refer to collocation relationship between words in a sentence. If a word can be followed by many different words, it will have many out-links; if a word follows many different words, it will have many in-links. If there are some pivot schemas in the children, such as "more ___", or "__ gone" (Tomasello, 1997), the pivot words may appear as the highly-linked nodes. But these words will not necessarily be hubs and authorities, as we discussed above. The positions of hubs and authorities are more intriguing, since they are determined by the global structure of the network, and not by the characteristics of individual nodes in isolation. The change of hubs and authorities may reflect the change of the network structure, which in turn reflects the change of the language system in the speaker.

It is expected that as adults' speech style is relatively stable, the structure of their lexical networks should be quite stable within speaker, and therefore the hubs and authorities should be largely consistent at different time instants. By contrast, during the process of development, the hubs and authorities in children's networks should vary, and be different from, but gradually approach to, those in the adults' networks. As to be shown in the following section, we find that the results from the analyses of children' and adult's networks support our expectation to a large extent. We also observe individual differences among children.

**4.2 Results of hub-authority analyses on adults' networks**

We first perform a hub-authority analysis on the stage networks of Joel's and Ruth's mothers. For each network, we pick out the ten most important hubs and ten most important authorities, as shown in Tables 2-5 (The number 10 is chosen arbitrarily for the sake of simplicity). The words are ranked according to their weights in a descending order. From the tables, we observe a high consistency both within each mother through the stages and across the two mothers. Especially, those with higher weights, such as the five most important hubs and authorities, rarely change. For example, "you", "it", "that", and "and" are always among the five most important hubs, while "a", "the", "you", "it", "that", "your", "in" are always important authorities. It is noted that these words all belong to the closed class, such as pronouns and propositions, instead of content words.

Table 2. Hubs in Joel's mother's stage networks.

| Stage 1 | Stage 2 | Stage 3 | Early Stage 4 | Late Stage 4 |
|---------|---------|---------|---------------|--------------|
| you | you | you | you | you |
| it | that | that | that | that |

| that | and | it | and | it |
|---|---|---|---|---|
| and | it | and | it | and |
| to | to | to | to | do |
| oh | like | got | got | got |
| do | not | he | like | like |
| is | do | it's | it's | not |
| what | got | that's | have | to |
| all | a | do | do | get |

Table 3. Authorities in Joel's mother's stage networks.

| Stage 1 | Stage 2 | Stage 3 | Early Stage 4 | Late Stage 4 |
|---|---|---|---|---|
| the | a | a | that | you |
| you | the | the | the | a |
| a | it | that | a | it |
| it | you | it | you | the |
| your | that | you | it | that |
| on | your | on | your | all |
| that | all | your | all | in |
| in | on | in | on | your |
| there | to | all | to | what |
| look | in | to | in | then |

Table 4. Hubs in Ruth's mother's stage networks.

| Stage 1 | Stage 2 | Early stage 3 | Last stage 3 | Stage 4 | Stage 5 |
|---|---|---|---|---|---|
| you | you | you | you | you | you |
| it | it | that | and | and | and |
| that | and | and | that | it | it |
| oh | to | oh | to | that | that |
| and | that | it | it | not | oh |
| to | not | to | on | oh | to |
| on | oh | like | is | well | not |
| she | do | have | get | is | like |
| is | on | is | not | to | I |
| this | that's | see | he | do | is |

Table 5. Authorities in Ruth's mother's stage networks.

| Stage 1 | Stage 2 | Early stage 3 | Last stage 3 | Stage 4 | Stage 5 |
|---|---|---|---|---|---|
| a | the | a | the | the | a |
| the | you | the | a | a | the |
| in | a | you | that | you | you |
| it | it | it | you | it | your |
| your | that | that | it | that | that |
| you | in | to | your | your | it |
| her | on | this | this | this | this |
| this | your | your | to | to | in |
| that | this | in | in | all | what |
| to | to | now | on | in | to |

To further illustrate the disassociation between degree and importance discussed earlier, we go on to pick out the nodes with the highest in- and out-degrees in the mothers' networks for comparison. As there is high consistency among mothers' networks across different stages, we only show in Table 6 the results of the last stage networks for Joel's and Ruth's mothers.

Comparing Table 6 with Tables 2-3 for Joel's mother, we can see that some nodes with the largest in-degrees, such as "and", "on", and "now", are not the most important authorities; some nodes with the largest out-degrees, such as "a", "the", "your", "he", and "it's", are not the best hubs. At the same time, words such as "do", "got" "like", "not" and "get", which do not have the highest number of out-degrees, appear as the best hubs. And words with only a medium number of in-degrees, such as "all", "what" and "then", are the best authorities. Ruth's mother's network exhibits a similar disparity between degree and importance, and even shares the same words with Joel's mother. For instance, besides "a" and "the", "your" has a large of out-degree, but it is not a good hub; "and" has a large in-degree, but not a good authority. In fact, there is obviously a large degree of consistency between the two mothers. Furthermore, the words with the highest degree are all in the closed-class vocabulary. This is consistent with what Ferrer & Solé (2001) have shown in their network constructed from a different source of data - a large contemporary corpus.

Table 6. Nodes with the highest in-degrees and out-degrees in Joel's and Ruth's mothers' last stage network.

| nodes with the 10 highest in-degree (Joel's mother) | nodes with the 10 highest out-degree (Joel's mother) | nodes with the 10 highest in-degree (Ruth's mother) | nodes with the 10 highest out-degree (Ruth's mother) |
|---|---|---|---|
| you | you | you | a |
| it | a | it | you |
| a | the | a | the |
| in | and | the | your |
| the | that | in | and |
| and | it | that | that |
| on | your | to | it |
| that | to | your | to |
| now | he | on | i |
| your | it's | and | not |

### 4.3 Results of Hub-authority analyses on children's networks

Tables 7-10 show the best 10 hubs and authorities in Joel's and Ruth's networks. As expected, in contrast to the stability exhibited in the mothers' lists, there is little consistency among the different stages for each child. In both children's networks, in the first stage there are many content words as hubs and authorities, such as "car" and "whale" in Joel's network and "mama", "baba" and "juice" in Ruth's network.

Table 7. Hubs in Joel's stage networks.

| Stage 1 | Stage 2 | Stage 3 | Early Stage 4 | Late Stage 4 |
|---|---|---|---|---|
| a | that | a | and | and |
| more | a | that's | a | got |
| two | like | it's | it's | you |
| the | not | got | got | want |
| mummy | and | like | you | that's |
| in | that's | want | i | is |
| that | want | the | he's | it's |
| on | the | on | that | in |
| there's | on | in | have | like |
| oh | in | put | want | was |

Table 8. Authorities in Joel's stage networks.

| Stage 1 | Stage 2 | Stage 3 | Early Stage 4 | Late Stage 4 |
|---------|---------|---------|---------------|--------------|
| car | it | that | a | it |
| one | there | my | it | a |
| daddy | that | a | my | that |
| pear | a | it | in | the |
| whale | the | the | the | you |
| chair | green | you | that | this |
| four | panda | there | go | my |
| two | car | one | you | in |
| there | In | in | this | me |
| hair | lettuce | to | some | there |

Table 9. Hubs in Ruth's stage networks.

| Stage 1 | Stage 2 | Early stage 3 | Last stage 3 | Stage 4 | Stage 5 |
|---------|---------|---------------|--------------|---------|---------|
| oh | me | me | me | that | you |
| no | no | mama | there | not | and |
| mama | mama | there | that | me | me |
| I | there | baba | baby | and | my |
| baba | more | want | want | want | going |
| me | baba | brick | no | baby | put |
| more | oh | no | mama | you | that |
| and | dada | oh | mummy | a | it |
| on | I | anna | the | get | got |
| there | and | more | a | got | want |

Table 10. Authorities in Ruth's stage networks.

| Stage 1 | Stage 2 | Early stage 3 | Last stage 3 | Stage 4 | Stage 5 |
|---------|---------|---------------|--------------|---------|---------|
| there | on | there | baby | that | a |
| on | there | me | me | me | you |
| baba | in | on | on | mummy | my |
| in | baba | mama | there | a | the |
| juice | mama | in | tea | my | little |
| mama | mmhm | baba | too | one | be |
| tea | anna | brick | tiger | the | in |
| fall | no | anna | one | go | it |
| no | me | sleep | mama | in | one |
| go | hiya | fall | back | there | go |

In later stages these content words gradually give way to function words. Closed-class words such as "you", "it", "that", and "and", which are very important nodes in adults' networks, enter the top 10 hub and authority lists at different stages, and secure their places thereafter (sometimes with variations like "it's", "that's"). One interesting discrepancy between the children and the mothers' list is that some authorities in mothers' networks, such as "a", "the", and "in", appear as hubs in the children's list, but never in mothers' list. We will discuss this in more detail in the next section.

Therefore, as reflected by the changes of the lists of the top 10 hub and authority nodes, children's lexical networks start off with few similarities with those of their mothers', but approach them gradually over time, acquiring more adult-like hubs and authorities.

## 4.4 The shift of articles in children's stage networks

We observe one interesting shift of the position of the two articles "a" and "the" in the hub and authority analyses. Intuitively, "a" and "the" should have many out-links, and have the potential to be good hubs. However, in adults' networks, we never find them in the first-10-best-hub lists, instead they consistently appear among the 10 best authorities. In comparison, in the children's networks, the two articles often appear as good hubs, but later disappear from the hub list, and appear in the authority list. The shifting process of the two articles may be taken as an indicator that the children's networks are approaching to the adult's. However, different children approach along different routes and at different rates.

To show the change of the two words, Table 11 lists their presence or absence in hub and authority lists across the stage networks of the four children we analyze earlier. In the table, 'Hub' means the word appears in the top 10 hub list and 'Authority' means it is in the top 10 authority list; '--' means it is in neither of the lists and 'H&A' means it is in both lists.

Table 11. The presence and absence of the two articles "a" and "the" in the lists of the first 10 best hubs and authorities in the four children's networks.

| **Joel** | stage 1 | stage 2 | stage 3 | early stage 4 | late stage 4 | |
|---|---|---|---|---|---|---|
| "a" | Hub | H & A | H & A | H & A | Authority | |
| "the" | Hub | H & A | H & A | Authority | Authority | |
| **Ruth** | stage 1 | stage 2 | early stage 3 | late stage 3 | stage 4 | stage 5 |
| "a" | -- | -- | -- | Hub | H & A | Authority |
| "the" | -- | -- | -- | Hub | Authority | Authority |
| **Carl** | early stage 3 | late stage 3 | stage 4 | early stage 5 | late stage 5 | |
| "a" | H & A | A | -- | A | -- | |
| "the" | H & A | A | -- | A | -- | |
| **Anne** | stage 2 | stage 3 | early stage 4 | middle stage 4 | late stage 4 | |
| "a" | -- | Authority | H & A | Authority | Authority | |
| "the" | -- | Authority | H & A | Authority | Authority | |

In Joel's networks, articles "a" and "the" both set out as hubs and later on switch to authorities gradually. There is a long intermediate period during the shifting process. By contrast, in Ruth's networks, neither of the two articles appears as important nodes until late stage 3, where they both are in the hub list. Then the two articles' positions change to authorities swiftly, and the whole shifting process completes within three stages, i.e. from late stage three to stage five.

To account for the differences of the different developmental trajectories between Joel and Ruth, we zoom in on the early networks of the two children. Figures 11 and 12 give a simple illustration of how the two children's starting points differ by showing the egonets of "a" and "the" in their stage-1 networks ("egonet" is a subset of the network which only includes the target node and its immediate neighbors).

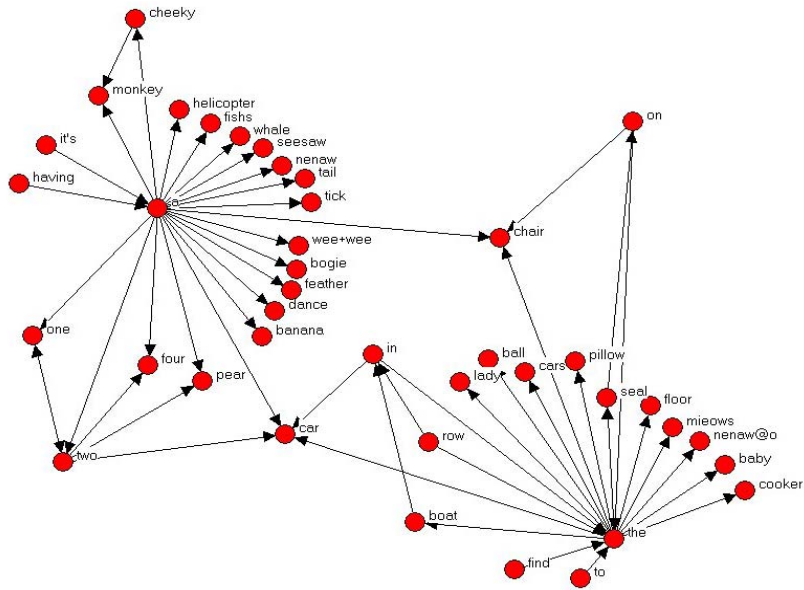

Figure 11. Egonet of "a" and "the" in Joel's stage-1 network.

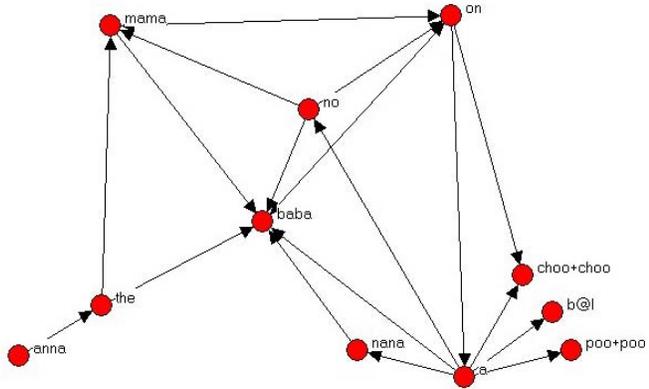

Figure 12. Egonet of "a" and "the" in Ruth's stage-1 network.

In Figure 11, articles "a" and "the" both connect to a large number of nouns, comprising noun phrases like "a car", "a chair", "the car", "the chair", etc. However, they seem to connect to two sets of nouns, with little overlap or links among each other. This is consistent with what other empirical studies have shown (Pine & Martindale, 1996). In Figure 12, there are much fewer noun phrases, and some of the nouns are child forms like "choo+choo', "poo+poo", etc. Therefore, it is not surprising that noun phrases like "a+NOUN" and "the + NOUN" don't play important roles in Ruth's networks at early stages.

Since stage 1 corresponds to an MLU range of [1,1.5], almost all the utterances in this stage are one or two words long and hence it is safe to assume that each link in the network is representative of a two-word utterance. In that sense, the differences shown in the two egonets above illustrate well the different two-word-utterance patterns of the two children. Joel's production is largely dominated by noun phrases but Ruth's is mostly expressive terms. These two children exemplify a traditional dichotomy of learning styles, that is, referential vs. expressive (Nelson, 1973).

The shift in the other two children, Carl and Anne, is not as clear and clean as in Joel and Ruth. For Carl, the two articles first appear as both a hub and an authority, and shift to become only authorities at stage 2. But in the subsequent stages the two words show fluctuation in their appearance in the authority list. It is expected they will stay as authorities in later stages. If subsequent data of the child are available, it would be interesting to test this prediction. For Anne, the articles are already authorities at the beginning of the investigation, and remain the same for the whole period. We suspect that the shifting process in Anne should have happened earlier before the study was carried out. The words appear temporarily as hubs also in Anne's early stage 4. The reason for this fluctuation requires further investigation in detail on Anne's data.

In both Joel and Ruth's shifting process, "the" appears as an authority earlier than "a". Based on this, we could infer that the children acquire the correct usage of "the" earlier than "a". However, this inference may not be valid. Empirical research on the acquisition of English articles has shown that children tend to overuse the definite article "the" when they have a referent they wish to introduce to someone for whom it is totally new in the discourse context. For example, with no introductory comments whatsoever they might tell a friend "Tomorrow we'll buy the toy" (Tomasello, 2003:210). The earlier shift of "the" may be due to this kind of overuse. More in-depth examination of the data is necessary to clarity this issue.

## 5. Conclusions and general discussions

In this study, we have proposed a new approach from a network perspective to examine language development. The network model allows us to measure children's incremental development by exploring the data available in existing corpus in a controlled manner. A number of new measures on the properties of the networks, such as size, connectivity, hub and authority analyses, etc., allow us to make quantitative comparison so as to reveal different paths of development. For example, the asynchrony of development in network size and average degree suggests that children cannot be simply classified as early talkers or late talkers by one or two measures. Children follow different paths in a multi-dimensional space. They may develop faster in one dimension but slower in another dimension.

In this usage-based study, we make few assumptions about how the data should be organized and interpreted in the network. The words are treated equally as nodes and collocations as links, without preprocessing of words and analyses on sentence structures. The characteristics of words and their usage emerge from the network automatically and are independent of any grammatical presumptions. The global patterns in the lexical networks can reveal developmental features which are not easily detected in traditional text analyses. In particular, the syntactical development may be revealed by changes of the global structure of the lexical networks. The analyses on hubs and authorities serve as a good example. The measure provides an alternative way to compare quantitatively children's languages with their care-takers'. In the case of the two articles "a" and "the", while they appear constantly as authorities in the adults' networks, we can see clearly that these two words exhibit a shifting process from hubs to authorities in most children's networks. And we observe different paths for this shifting process in the children.

The measures proposed in this paper, such as the network size, connectivity and the importance of nodes, are only a few examples of simple analyses for networks, to trace the development of the children's language system. There are many other possible analyses to perform on the lexical networks, such as degree distribution, the correlation between degree and. clustering coefficient (to detect the presence of hierarchical structures), centrality distribution, assortativeness / disassortativeness, and so on (Ferrer et al, 2004). These analyses may provide more quantitative measures for language development, which have not been explored by traditional methods.

While the new measures have provided some new ways to evaluate the development and have suggested some interesting characteristics, however, there is still lack of answers how to interpret the meaning of these new measures, for example, the hubs and authorities in the network. Though the articles shifting to be authorities may signify the attainment of a certain grammatical knowledge, it will be more satisfactory if deeper linguistic interpretation for the hubs and authorities can be found.

Moreover, the collocation relation used in this study only provides a very simple way to construct networks. The immediate challenge is to represent more complex linguistic structures in the network model, such as the non-adjacent constructions. For example, in the sentence "the man who likes to wear sun glasses in the restaurant is a movie star", the verb "is" depends on the noun phrase "the man", which is located far ahead in the beginning of the sentence. Non-adjacent dependency is a central characteristic of human language. There are two possible ways to take into account such constructions in the network models. One is to use corpora which have been tagged with syntactic parsing (Ferrer et al, 2004), but such corpora are much fewer available for use, and they are prone to one serious problem as mentioned earlier, which is the assumption that the children have the same grammatical knowledge as the adult speakers (the analyst who tags the children's speech data). The other solution is to adopt more complex representations of the nodes and the links, such as adding semantic encoding to the nodes, and using weighted links to allow long distance collocation to be maintained. So far the networks are not weighted, i.e. the links are all of the same importance. However, in actual speech data, some collocations and constructions are more frequent than others, and children are shown to be good statisticians to detect such distributional and frequency features from very early on (Saffran et al., 1996). If the collocation appears frequently, they may be acquired as fixed constructions. The effect of frequency in language change, language processing and language acquisition have been received more attention in recent years (Bybee & Hopper, 2001; Ellis, 2002). Future models of lexical networks should take into account the frequency of occurrence in the data.

The networks in this study, constructed based on collocation relationships, are mainly aimed for revealing the development of syntactic structures. However, the words can be built up as networks according to their relationships in meanings (Motter et al., 2002; Sigman & Cecchi, 2002). Development of the semantic networks is another fruitful area to explore, to see how children's conceptual networks grow in different ways.

Language acquisition presents a rich area for network theories to explore. On the one hand, huge amount of data sets are available for different languages and for both first and second language acquisition (such as those available in CHILDES); the network analyses can be applied on these data to reexamine many old questions in acquisition, and suggest new measures to evaluate development quantitatively. On the other hand, the need to analyze the data may raise new questions for network theory, including network construction and analyses. In particular, the growth of linguistic networks during the language acquisition process poses a good challenge for network growth modeling, which has been an area with increasing importance in network research (EurPhy, 2004). Therefore, to apply a network perspective in studying language acquisition will bring a mutual benefit to both areas.

**Acknowledgements:**

We would like to thank Professors William W-Y Wang, Thomas H-T Lee, Ron G-R Chen, and the members in the former Language Engineering Laboratory in City University of Hong Kong, and the Laboratoire Dynamique du Langage in Lyon, France for their support and advice for this study. Also, we thank Professors Nick Ellis, Ricard Solé and Dr. Christophe Coupé for their valuable comments on the paper.